\PassOptionsToPackage{margin=1in}{geometry}
\documentclass[12pt]{article}
\usepackage{wordlike}
\usepackage{setspace}
\usepackage{booktabs} 
\usepackage{tabularx}
\usepackage{array}
\usepackage{graphicx}
\usepackage{multirow}
\usepackage{appendix}
\usepackage{svg}
\usepackage{float}
\usepackage{newtxtext,newtxmath}
\usepackage{adjustbox}
\usepackage{titlesec}
\usepackage{titling}

\posttitle{\par\end{center}}

\titleformat{\section}[block]
{\Large\bfseries}%
{\thesection}
{5mm}
{}
\titlespacing*{\section}{2pt}{*2}{2pt}

\titleformat{\subsection}[block]
{\large\bfseries}%
{\thesubsection}
{5mm}
{}

\titleformat{\subsubsection}[block]
{\large\bfseries}%
{\thesubsubsection}
{5mm}
{}

\begin{document}

\title{A Deep Learning Architecture for De-identification of Patient Notes:
  Implementation and Evaluation}  

\author{Kaung Khin$^1$ \and Philipp Burckhardt$^2$ \and Rema Padman$^3$}
\date{%
  {kkhin$^1$, pburckhardt$^2$, rpadman$^3$}@cmu.edu\\
  Carnegie Mellon University\\
    Pittsburgh, PA, USA
}

\maketitle

\section{Introduction}
\label{sec:intro}
\footnote{A version of this paper was submitted to the 28th Workshop on Information Technologies and Systems}
Electronic Health Records (EHR) have become ubiquitous in recent years in the
United States, owing much to the The Health Information Technology
for Economic and Clinical Health (HITECH) Act of 2009. \cite{EMR_adoption} Their
ubiquity have given researchers a treasure trove of new data, especially in the
realm of unstructured textual data. However, this new data source comes with usage
restrictions in order to preserve the privacy of individual patients as mandated by
the Health Insurance Portability and Accountability Act (HIPAA). HIPAA
demands any researcher using this sensitive data to first strip the medical
records of any protected health information (PHI), a process known as de-identification.

HIPAA allows for two methods for de-identifying PHIs: the ``Expert
Determination'' method in which an expert certifies that the information is
rendered not individually identifiable, and the ``Safe Harbor'' method in which
18 identifiers are removed or replaced with random data in order for the data to be
considered  not individually identifiable. Our research pertains
to the second method (a list of the relevant identifiers can be seen in Table \ref{tab:PHITypes}).

The process of de-identification has been largely a manual and labor intensive
task due to both the sensitive nature of the data and the limited availability
of software to automate the task. This has led to a relatively small number of open
health data sets available for public use. Recently, there have been two well-known
de-identification challenges organized by Informatics for Integrating
Biology and the Bedside (i2b2) to encourage innovation in the field of de-identification.

In this paper, we build on the recent advances in natural language processing, 
especially with regards to word embeddings, by incorporating deep contextualized
word embeddings developed by Peters et al. \cite{elmo_embeddings} into a deep
learning architecture. More precisely, we present a deep learning architecture
that differs from current architectures in literature by using bi-directional
long short-term memory networks (Bi-LSTMs) with variational dropouts and
deep contextualized word embeddings while also using components already present
in other systems such traditional word embeddings, character LSTM embeddings and
conditional random fields. We test this architecture on two gold standard data sets, the 2014 i2b2
de-identification Track 1 data set \cite{i2b2_review} and the nursing notes corpus
\cite{neamatullah08_autom_de_ident_free_text_medic_recor}. The architecture
achieves state-of-the-art performance on both data sets while also achieving faster convergence without the use of
dictionaries (or gazetteers) or other rule-based methods that are typically used
in other de-identification systems. 

The paper is organized as follows: In Section \ref{sec:relatedWork}, we review the latest
literature around techniques for de-identification with an emphasis on related
work using deep learning techniques. In Section \ref{sec:method}, we detail
our deep learning architecture and also describe how we use the deep
contextualized word embeddings method to improve our
results. Section \ref{sec:data} describes the two data sets we will use to evaluate
our method and our evaluation metrics. Section
\ref{sec:evalAndResults} presents the performance of our architecture on the
data sets. In Section \ref{sec:disAndFurther}, we discuss the results and provide
an analysis of the errors. Finally, in Section \ref{sec:conclusion}, we summarize
our contributions while also discussing possible future research. 

\begin{table}[h]
\centering
  \caption{Protected Health Information Types \cite{hhs}}
  \label{tab:PHITypes}
\begin{tabular}{cl}
  \toprule
  \textbf{No.} & \textbf{PHI Type}                                                                                                                   \\ 
  \midrule
  1   & Names \\
  2   & All geographic subdivisions smaller than a state \\
  3   & Dates \\
  4   & Telephone Numbers \\
  5   & Vehicle Identifiers \\
  6   & Fax Numbers \\
  7   & Device Identifiers and Serial Numbers \\
  8   & Emails \\
  9   & URLs \\
  10  & Social Security Numbers \\
  11  & Medical Record Numbers \\
  12  & IP Addresses \\
  13  & Biometric Identifiers \\
  14  & Health Plan Beneficiary Numbers \\
  15  & Full-face photographic images and any comparable images \\
  16  & Account Numbers \\
  17  & Certificate/license numbers \\
  18  & Any other unique identifying number, characteristic, or code. \\
  \bottomrule
\end{tabular}
\end{table}

\section{Background and Related Work}
\label{sec:relatedWork}

The task of automatic de-identification has been heavily studied recently, in
part due to two main challenges organized by i2b2 in 2006 and in 2014.
The task of de-identification can be classified as a named entity recognition (NER) problem
which has been extensively studied in machine learning literature. Automated
de-identification systems can be roughly broken down into four main categories:
\begin{itemize}
\item Rule-based Systems
\item Machine Learning Systems
\item Hybrid Systems
\item Deep Learning Systems \footnote{Deep learning is technically a subset of machine learning}
\end{itemize}

\subsection{Rule-based Systems}
\label{sec:rulesystems}

Rule-based systems make heavy use of pattern matching such as
dictionaries (or gazetteers), regular expressions and other patterns. \cite{i2b2_review} Systems such as the
ones described in \cite{morrison09_repur_clinic_recor,
  uzuner07_evaluat_state_of_art_autom_de_ident} do not require the use any
labeled data. Hence, they are considered as unsupervised learning systems. 
Advantages of such systems include their ease of use, ease of adding new patterns and
easy interpretability. However, these methods suffer from a lack of robustness
with regards to the input. For example, different casings of the same word could
be misinterpreted as an unknown word. Furthermore, typographical errors are
almost always present in most documents and rule-based systems often cannot
correctly handle these types of inaccuracies present in the data. Critically, these
systems cannot handle context which could render a medical text unreadable. For
example, a diagnosis of ``Lou Gehring disease'' could be misidentified by such a
system as a PHI of type \textit{Name}. The system might replace the tokens ``Lou'' and ``Gehring'' with
randomized names rendering the text meaningless if enough of these tokens were replaced.

\subsection{Machine Learning Systems}
\label{sec:mlsystems}

The drawbacks of such rule-based systems led researchers to adopt a machine learning
approach. A comprehensive review
of such systems can be found in \cite{ferrandez12_evaluat_curren_autom_de_ident,
  deidReview}. In machine learning systems, given a sequence of input vectors $\mathbf{X}_1 \cdots
\mathbf{X}_n$, a machine learning algorithm outputs label predictions $\mathbf{Y}_1 \cdots \mathbf{Y}_n$.
Since the task of de-identification is a classification task, traditional
classification algorithms such as support vector machines, conditional random fields (CRFs) and decision trees
\cite{MIT_dernoncourt} have been used for building de-identification systems.

These machine learning-based systems have the advantage of
being able to recognize complex patterns that are not as readily evident to the naked
eye. However, the drawback of such ML-based systems is that since classification
is a supervised learning task, most of the common classification algorithms
require a large labeled data set for robust models. Furthermore, since most of the algorithms
described in the last paragraph maximize the likelihood of a label $\mathbf{Y}_i$ given
an vector of inputs $\mathbf{X}_i$, rare patterns that might not occur in the training
data set would be misclassified as not being a PHI label. Furthermore, these models
might not be generalizable to other text corpora that contain significantly
different patterns such as sentence structures and uses of different abbreviated
words found commonly in medical notes than the training data set.

\subsection{Hybrid Systems}
\label{sec:hybridsystems}

With both the advantages and disadvantages of stand alone rule-based and
ML-based systems well-documented, systems such as the ones detailed in
\cite{i2b2_review} combined both ML and rule-based systems to
achieve impressive results. Systems such as the ones presented for 2014 i2b2
challenge by Yang et al. \cite{nottingham_i2b2_paper} and Liu et
al. \cite{liu15_autom_de_ident_elect_medic} used dictionary look-ups, regular expressions and
CRFs to achieve accuracies of well over 90\% in identifying PHIs.

It is important to note that such hybrid systems rely heavily on feature
engineering, a process that manufactures new features from the data that are not present
in the raw text. Most machine learning techniques, for example,
cannot take text as an input. They require the text to be represented as a vector
of numbers. An example of such features can be seen in the system that won the
2014 i2b2 de-identification challenge by Yang et al.
\cite{nottingham_i2b2_paper}. Their system uses token features such as
part-of-speech tagging and chunking, contextual features such as word lemma and
POS tags of neighboring words, orthographic features such as capitalization and
punctuation marks and task-specific features such as building a list that
included all the full names, acronyms of US states and collecting
TF-IDF-statistics. Although such hybrid systems achieve impressive results, the task of feature
engineering is a time-intensive task that might not be generalizable to other text corpora.  

\subsection{Deep Learning Systems}
\label{sec:dlsys}

With the disadvantages of the past three approaches to building a
de-identification system in mind, the current state-of-the-art systems employ
deep learning techniques to achieve better results than the best hybrid systems
while also not requiring the time-consuming process of feature engineering. Deep
learning is a subset of machine learning that uses multiple layers of Artificial Neural
Networks (ANNs), which has been very succesful  at most Natural Language Processing (NLP) tasks. Recent advances in the
field of deep learning and NLP especially in regards to named entity recognition
have allowed systems such as the one by Dernoncourt et al.
\cite{MIT_dernoncourt} to achieve better results on the 2014 i2b2
de-identification challenge data set than the winning hybrid system proposed by
Yang et al. \cite{nottingham_i2b2_paper}. The advances in NLP and deep learning
which have allowed for this performance are detailed below.

\subsubsection{Embeddings}
\label{sec:embed}

ANNs cannot take words as inputs and require numeric inputs, therefore, past approaches to using ANNs for
NLP have been to employ a bag-of-words (BoW) representation of words where a
dictionary is built of all known words and each word in a sentence is
assigned a unique vector that is inputted into the ANN. A drawback of such a
technique is such that words that have similar meanings are represented completely
different. As a solution to this problem, a technique called word embeddings have been used.
Word embeddings gained popularity when Mikolov et al. \cite{glove} used ANNs to generate a
distributed vector representation of a word based on the usage of
the word in a text corpus. This way of representing words allowed for similar words
to be represented using vectors of similar values while also allowing for complex operations
such as the famous example: $X_{King} - X_{Man} + X_{Woman} = X_{Queen}$, where
$X$ represents a vector for a particular word.

While pre-trained word embeddings such as the widely used \textit{GloVe} \cite{glove}
embeddings are revolutionary and powerful, such representations only capture one
context representation, namely the one of the training corpus they were derived from. This
shortcoming has led to the very recent development of context-dependent
representations such as the ones developed by \cite{elmo_embeddings,
  mccann17_learn_trans}, which can capture different features of a word.

The Embeddings from Language Models (ELMo) from the system by Peters et al.
\cite{elmo_embeddings} are used by the architecture in this paper to achieve
state-of-the-art results. The \textit{ELMo} representations, learned by combining
Bi-LSTMs with a language modeling objective, captures context-depended aspects
at the higher-level LSTM while the lower-level LSTM captures aspects of syntax.
Moreover, the outputs of the different layers of the system can be used independently
or averaged to output embeddings that significantly improve some existing models
for solving NLP problems. These results drive our motivation to include the
\textit{ELMo} representations in our architecture.

\subsubsection{Neural Networks}
\label{sec:nn}

The use of ANNs for many machine learning tasks has gained popularity in recent
years.  Recently, a variant of recurrent neural networks (RNN) called
Bi-directional Long Short-Term Memory (Bi-LSTM) networks has been successfully
employed especially in the realm of NER.

In fact, several Bi-LSTM architectures have been proposed to tackle the problem of NER:
LSTM-CRF, LSTM-CNNs-CRF and LSTM-CNNs \cite{MIT_dernoncourt}. The current best
performing system on the i2b2 dataset is in fact a system based on LSTM-CRF
\cite{MIT_dernoncourt}.

\section{Method}
\label{sec:method}

\begin{figure}[h]
\caption{Deep Learning Architecture}
\label{fig:nnarchi}
\includegraphics[width=\textwidth]{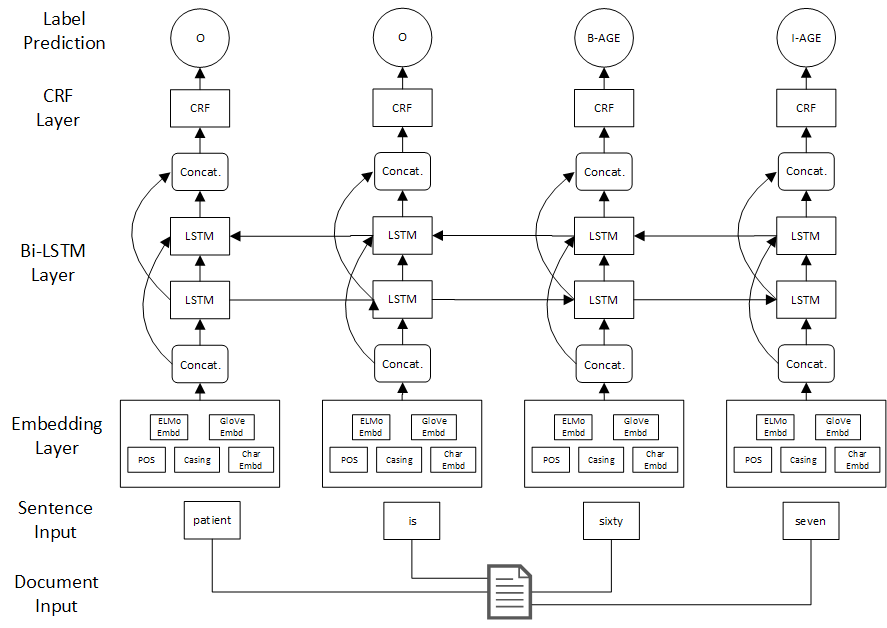}
\end{figure}

Our architecture incorporates most of the recent advances in NLP and NER while also
differing from other architectures described in the previous section by use of
deep contextualized word embeddings, Bi-LSTMs with a variational dropout and the
use of the Adam optimizer. Our architecture can be broken down into four
distinct layers: pre-processing, embeddings, Bi-LSTM and CRF classifier. A
graphical illustration of the architecture can be seen in Figure
\ref{fig:nnarchi} while a summary of the parameters for our architecture can be found in Table \ref{tab:paramsum}.

\begin{table}
\centering
\caption{Summary of architecture parameters}
\label{tab:paramsum}
\begin{tabular}{|lcc|} 
\hline
\multicolumn{1}{|c}{\textbf{Type}} & \textbf{Details}     & \textbf{Parameters}  \\ 
\hline
\textit{ELMo }embeddings           & Averaged Layers      & 1024                 \\
Token embedding                    & \textit{GloVe 3}     & 300                  \\
Part-of-speech embeddings          & Generated from nltk  & 20                   \\
Casing embeddings                  &                      & 20                   \\
Character LSTMs                    & Two LSTMs            & 25                   \\
Bi-LSTM                            & Two LSTMs            & 100                  \\
Variational Dropout                & \multicolumn{1}{l}{} & 0.5                  \\
Optimizer                          & \multicolumn{1}{l}{} & Adam                 \\
\hline
\end{tabular}
\end{table}

\subsection{Pre-processing Layer}
\label{sec:preprocess}

For a given document $\mathbf{d}_i$, we first break down the document into sentences
$\mathbf{s}_{i,j}$, tokens $\mathbf{t}_{i,j,k}$ and characters $\mathbf{c}_{i,j,k,l}$ where $i$
represents the document number, $j$ represents the sentence number, $k$
represents the token number, and $l$ represents the character number. For
example,  $\mathbf{t}_{1,2,3} = $ Patient, where the token: ``Patient'' represents the
3rd token of the 2nd sentence of the 1st document.

After parsing the tokens, we use a widely used and readily available Python toolkit called
Natural Langauge ToolKit (NLTK)  to generate a
part-of-speech (POS) tag for each token. This generates a POS
feature for each token which we will transform into a 20-dimensional
one-hot-encoded input vector, $\mathbf{POS}_{i,j,k}$, then feed into the main LSTM layer.

For the data labels, since the data labels can be made up of multiple tokens,
we formatted the labels to the BIO scheme. The BIO scheme tags the beginning of
a PHI with a \textit{B-}, the rest of the same PHI tokens as \textit{I-} and the
rest of the tokens not associated with a PHI as \textit{O}. For example, the
sentence, ``$\texttt{The patient is sixty seven years old}$'', would have the corresponding labels,
``$\texttt{O O O B-AGE I-AGE O O}$''. 

\subsection{Embedding Layer}
\label{sec:embedlayer}

For the embedding layer, we use three main types of embeddings to represent our
input text: traditional word embeddings, \textit{ELMo} embeddings and character-level
LSTM embeddings. 

The traditional word embeddings use the latest \textit{GloVe} 3 \cite{glove} pre-trained
word vectors that were trained on the Common Crawl with about 840 billion
tokens. For every token input, $\mathbf{t}_{i,j,k}$, the \textit{GloVe} system outputs $\mathbf{Glove}_{i,j,k}$, a dense
300-dimensional word vector representation of that same token. We also
experimented with other word embeddings by using the bio-medical corpus trained
word embeddings \cite{bioEmbeddingSource} to see if having word embeddings
trained on medical texts will have an impact on our results. 

As mentioned in previous sections, we also incorporate the powerful \textit{ELMo}
representations as a feature to our Bi-LSTMs. The specifics of the \textit{ELMo}
representations are detailed in \cite{elmo_embeddings}. In short, we compute an
\textit{ELMo} representation by passing a token input $\mathbf{t}_{i,j,k}$ to
the \textit{ELMo} network and averaging the the layers of the network to produce
an 1024-dimensional \textit{ELMo} vector, $\textbf{ELMo}_{i,j,k}$.

Character-level information can capture some information about the token itself
while also mitigating issues such as unseen words and misspellings. While
lemmatizing (i.e., the act of turning inflected forms of a word to their base or
dictionary form) of a token can solve these issues, tokens such as the ones found in
medical texts could have important distinctions between, for example, the grammar
form of the token. As such, Ma et al. \cite{ma16} have used Convolutional
Neural Networks (CNN) while Lample et al. \cite{nnArchiForNER} have used Bi-LSTMs
to produce character-enhanced representations of each unique token. We have
utilized the latter approach of using Bi-LSTMs for produce a character-enhanced
embedding for each unique word in our data set.\footnote{Words appearing in the
  test set that do not appear in the training set are mapped to the UNKNOWN
  representation} Our parameters for the forward and backward LSTMs are 25 each
and the maximum character length is 25, which results in an 50-dimensional embedding
vector, $\mathbf{CEmbd}_{i,j,k}$, for each token. 

After creating the three embeddings for each token, $\mathbf{t}_{i,j,k}$, we concatenate
the \textit{GloVe} and \textit{ELMo} representations to produce a single 1324-dimensional word input vector,
$\textbf{WInput}_{i,j,k}$. The concatenated word vector is then further concatenated
with the character embedding vector, $\mathbf{CEmbd}_{i,j,k}$, POS
one-hot-encoded vector, $\mathbf{POS}_{i,j,k}$, and the casing embedded vector,
$\mathbf{Casing}_{i,j,k}$, to produce a single 1394-dimensional input vector,
$\mathbf{Input}_{i,j,k}$, that we feed into our Bi-LSTM layer.

\subsection{Bi-LSTM Layer}
\label{sec:bilstmlayer}

The Bi-LSTM layer is composed of two LSTM layers, which are a variant of the
Bidirectional RNNs. In short, the Bi-LSTM layer contains
two independent LSTMs in which one network is fed input in the normal time
direction while the other network is fed input in the reverse time direction. The outputs of
the two networks can then be combined using either summation, multiplication,
concatenation or averaging. Our architecture uses simple concatenation to
combine the outputs of the two networks.

Our architecture for the Bi-LSTM layer is similar to the ones used by
\cite{nnArchiForNER, neuroNER, harbin_deid_using_rnn_and_crf} with each LSTM
containing 100 hidden units. To ensure that the neural networks do not
overfit, we use a variant of the popular dropout technique
called variational dropout \cite{vardropout} to regularize our neural networks.
Variational dropout differs from the traditional na\"ive dropout technique by
having the same dropout mask for the inputs, outputs and the recurrent layers
\cite{vardropout}. This is in contrast to the traditional technique of applying
a different dropout mask for each of the input and output layers.
\cite{reimers17_optim_hyper_deep_lstm_networ} shows that variational dropout
applied to the output and recurrent units performs significantly better
than na\"ive dropout or no dropout for the NER tasks. As such, we apply
a dropout probability of 0.5 for both the output and the recurrent units in our
architecture. 

\subsection{CRF layer}
\label{sec:crflayer}

As a final step, the outputs of the Bi-LSTM layer are inputted into a linear-chain CRF
classifier, which maximizes the label probabilities of the entire input sentence. This
approach is identical to the Bi-LSTM-CRF model by Huang et al. \cite{huang15_bidir}
CRFs have been incorporated in numerous state-of-the-art models
\cite{nnArchiForNER, harbin_deid_using_rnn_and_crf,
  neamatullah08_autom_de_ident_free_text_medic_recor} because of their ability to
incorporate tag information at the sentence level. 

While the Bi-LSTM layer takes information from the context into account when
generating its label predictions, each decision is independent from the other
labels in the sentence. The CRF allows us to find the labeling sequence in a
sentence with the highest probability. This way, both previous and subsequent
label information is used in determining the label of a given token. As a
sequence model, the CRF posits a probability model for the label sequence of the
tokens in a sentence, conditional on the word sequence and the output scores
from the Bi-LTSM model for the given sentence. In doing so, the CRF models the
conditional distribution of the label sequence instead of a joint distribution
with the words and output scores. Thus, it does not assume independent features,
while at the same time not making strong distributional assumptions about the
relationship between the features and sequence labels.

\section{Data and Evaluation Metrics}
\label{sec:data}

The two main data sets that we will use to evaluate our architecture are the 2014
i2b2 de-identification challenge data set \cite{i2b2_review} and the nursing
notes corpus \cite{neamatullah08_autom_de_ident_free_text_medic_recor}.

The i2b2 corpus was used by all tracks of the 2014 i2b2 challenge. It consists
of 1,304 patient progress notes for 296 diabetic patients. All the PHIs were
removed and replaced with random replacements. The PHIs in this data set were
broken down first into the HIPAA categories and then into the i2b2-PHI categories as shown
in Table \ref{tab:phibreakdowni2b2}. Overall, the data set contains 56,348 sentences with
984,723 separate tokens of which 41,355 are separate PHI tokens, which
represent 28,867 separate PHI instances. For our test-train-valid split, we chose
10\% of the training sentences to serve as our validation set, which represents
3,381 sentences while a separately held-out official test data set was specified by the
competition. This test data set contains 22,541 sentences including 15,275 separate PHI tokens.

The nursing notes were originally collected by Neamatullah et al.
\cite{neamatullah08_autom_de_ident_free_text_medic_recor}. The data set contains 2,434 notes of which there are 1,724 separate PHI
instances. A summary of the breakdown of the PHI categories of this nursing
corpora can be seen in Table \ref{tab:phibreakdowni2b2}.

\begin{table}
\centering
\caption{PHI Categories Breakdown between HIPPA and i2b2}
\label{tab:phibreakdowni2b2}
\begin{tabular}{|cll|} 
\hline
HIPPA           & i2b2                                                                          & Nursing corpus         \\ 
\hline
Name            & Patient, Doctor,                                                              & Patient Name, Initial  \\
\multicolumn{1}{|l}{}    & Username                                                                      & Clinician, Proxy       \\
Profession      & Profession                                                                    & -                      \\
Location        & Street, City, State,                                                          & Location               \\
\multicolumn{1}{|l}{} & \begin{tabular}[c]{@{}l@{}}Country, Zip, Hospital,\\Organization\end{tabular} &                        \\
Age                   & Age                                                                           & -                      \\
Date                  & Date                                                                          & Date                   \\
Contact               & Phone, Fax, Email,                                                            & Phone                  \\
\multicolumn{1}{|l}{} & URL, IP Address                                                               &                        \\
ID                    & Medical Record, ID No,                                                        & -                      \\
\multicolumn{1}{|l}{} & SSN~, License No                                                               &                        \\
\hline
\end{tabular}
\end{table}

\subsection{Evaluation Metrics}
\label{sec:evalMetrics}

For de-identification tasks, the three metrics we will use to evaluate the
performance of our architecture are Precision, Recall and $F_1$ score as
defined below. We will compute both the binary $F_1$ score and the three
metrics for each PHI type for both data sets. Note that binary $F_1$ score
calculates whether or not a token was identified as a PHI as opposed to
correctly predicting the right PHI type. For de-identification, we place more
importance on identifying if a token was a PHI instance with correctly
predicting the right PHI type as a secondary objective. 

\[ \text{Precision} = \frac{\textnormal{No. of Correctly Identified PHI Tokens}}{\textnormal{No. of Tokens
    Identified as PHI Tokens}} \]

\[ \text{Recall} = \frac{\textnormal{No. of Correctly Identified PHI
      Tokens}}{\textnormal{Total No. of PHI Tokens}} \]

\[ F_1 = 2 * \frac{precision*recall}{precision + recall} \]

Notice that a high recall is paramount given the risk of accidentally disclosing
sensitive patient information if not all PHI are detected and removed from the
document or replaced by fake data. A high precision is also desired to preserve the
integrity of the documents, as a large number of false positives might obscure
the meaning of the text or even distort it. As the harmonic mean of precision
and recall, the $F_1$ score gives an overall measure for model performance that
is frequently employed in the NLP literature.

As a benchmark, we will use the results of the systems by Burckhardt et al.
\cite{philip_deidentify}, Liu et al. \cite{harbin_deid_using_rnn_and_crf}, 
Dernoncourt et al.\cite{MIT_dernoncourt} and Yang et al.
\cite{nottingham_i2b2_paper} on the i2b2 dataset and the
performance of Burckhardt et al. on the nursing corpus. Note that Burckhardt et
al. used the entire data set for their results as it is an unsupervised learning
system while we had to split our data set into 60\% training data and 40\% testing data. 

\section{Results}
\label{sec:evalAndResults}

We evaluated the architecture on both the i2b2-PHI categories and the HIPAA-PHI
categories for the i2b2 data set based on token-level labels. Note that the
HIPAA categories are a super set of the i2b2-PHI categories. We also ran the
analysis 5+ times to give us a range of maximum scores for the different data
sets. 

Table \ref{tab:i2b2binary} gives us a summary of how our architecture performed
against other systems on the binary $F_1$ score metrics while Table
\ref{tab:HIPAAphi} and Table \ref{tab:i2b2phiscores} summarizes the performance
of our architecture against other systems on HIPAA-PHI categories and i2b2-PHI categories respectively.
Table \ref{tab:nursingresults} presents a summary of the performance on the nursing note corpus 
while also contrasting the performances achieved by the \textit{deidentify} system. 

\begin{table}
\centering
\caption{Binary F-1 scores comparison}
\label{tab:i2b2binary}
\begin{tabular}{|c|c|c|c|} 
\hline
\multirow{2}{*}{Model} & \multicolumn{3}{c|}{i2b2 data set}  \\ 
\cline{2-4}
                       & Precision & Recall & $F_1$             \\ 
\hline
Our architecture                    & \textbf{0.9830}    & 0.9737 & 0.9783         \\
Dernoncourt et al. \cite{MIT_dernoncourt}   & 0.9792    & \textbf{0.9783} & \textbf{0.9787}         \\
Liu et al. \cite{harbin_deid_using_rnn_and_crf}           & 0.9646    & 0.9380 & 0.9511         \\
Yang et al. \cite{nottingham_i2b2_paper}  & 0.9815 & 0.9414 & 0.9611 \\
Burckhardt et al. \cite{philip_deidentify}   & 0.5989 & 0.8296 & 0.6957 \\
\hline
\end{tabular}
\end{table}

\begin{table}
\centering
\caption{HIPAA-PHI result}
\label{tab:HIPAAphi}
  \begin{adjustbox}{width=\textwidth}
\begin{tabular}{|c|ccc|ccc|ccc|} 
\hline
\multirow{2}{*}{HIPAA PHI} & \multicolumn{3}{c|}{Our architecture} & \multicolumn{3}{c|}{Liu et al.} & \multicolumn{3}{c|}{Yang et al.}  \\ 
\cline{2-10}
                           & Precision & Recall~ & $F_1$           & Precision & Recall & $F_1$      & Precision & Recall & $F_1$        \\ 
\hline
Name                       & 0.9832    & 0.9721  & \textbf{0.9776}          & 0.9542    & 0.9403 & 0.9472     & 0.9696    & 0.9168 & 0.9464       \\
Profession                 & 0.9099    & 0.8878  & \textbf{0.8987}          & 0.9134    & 0.6480 & 0.7582     & 0.8106    & 0.5978 & 0.6881       \\
Location                   & 0.9566    & 0.9427  & \textbf{0.9496}          & 0.9266    & 0.8500 & 0.8867     & 0.9061    & 0.7612 & 0.8273       \\
Age                        & 0.9694    & 0.9756  & 0.9725          & 0.9866    & 0.9634 & \textbf{0.9748}     & 0.9712    & 0.9254 & 0.9477       \\
Date                       & 0.9928    & 0.9821  & \textbf{0.9874}          & 0.9834    & 0.9541 & 0.9652     & 0.9867    & 0.9663 & 0.9764       \\
Contact                    & 0.9780    & 0.9834  & \textbf{0.9807}          & 0.9765    & 0.9541 & 0.9652     & 0.9663    & 0.9220 & 0.9437       \\
ID                         & 0.9238    & 0.9496  & \textbf{0.9365}          & 0.9441    & 0.9184 & 0.9311     & 0.9378    & 0.9168 & 0.9272       \\
\hline
\end{tabular}
\end{adjustbox}
\end{table}

\begin{table}
\centering
\caption{Scores broken down by i2b2 PHI categories}
\label{tab:i2b2phiscores}

\begin{tabular}{|c|ccc|ccc|} 
\hline
\multicolumn{1}{|c|}{\multirow{2}{*}{PHI Type}} & \multicolumn{3}{c|}{Our architecture}       & \multicolumn{3}{c|}{Yang et al.}      \\ 
\cline{2-7}
\multicolumn{1}{|c|}{}                          & Precision & Recall & $F_1$            & Precision & Recall & $F_1$            \\ 
\hline
Patient                                         & 0.9655    & 0.9555 & \textbf{0.9605 } & 0.9602    & 0.9067 & 0.9327           \\
Doctor                                          & 0.9679    & 0.9561 & \textbf{0.962 }  & 0.9723    & 0.9195 & 0.9452           \\
Username                                        & 0.9565    & 0.9565 & 0.9565           & 1.0       & 0.9665 & \textbf{0.9778}  \\
Profession                                      & 0.9099    & 0.8878 & \textbf{0.8987 } & 0.8106    & 0.5978 & 0.6881           \\
Country                                         & 0.8174    & 0.7923 & \textbf{0.8046 } & 0.7857    & 0.1888 & 0.3034           \\
State                                           & 0.9587    & 0.9073 & \textbf{0.9323 } & 0.9222    & 0.8105 & 0.8627           \\
City                                            & 0.877     & 0.9441 & \textbf{0.9093 } & 0.8598    & 0.7077 & 0.7764           \\
Street                                          & 0.9878    & 0.9902 & \textbf{0.989 }  & 0.9851    & 0.9706 & 0.9778           \\
Zip                                             & 0.9856    & 0.9785 & 0.982            & 1.0       & 0.9714 & \textbf{0.9855}  \\
Hospital                                        & 0.9257    & 0.9293 & \textbf{0.9275 } & 0.9009    & 0.8309 & 0.8644           \\
Organization                                    & 0.8518    & 0.6216 & \textbf{0.7187 } & 0.7143    & 0.3049 & 0.4274           \\
Age                                             & 0.9694    & 0.9756 & \textbf{0.9725 } & 0.9712    & 0.9254 & 0.9477           \\
Date                                            & 0.9928    & 0.9821 & \textbf{0.9874 } & 0.9867    & 0.9663 & 0.9764           \\
Phone                                           & 0.9583    & 0.9829 & \textbf{0.9704 } & 0.9803    & 0.9256 & 0.9522           \\
Medical Record                                  & 0.9661    & 0.9925 & \textbf{0.9791 } & 0.9649    & 0.9763 & 0.9706           \\
ID No.                                          & 0.7173    & 0.7983 & 0.7557           & 0.8681    & 0.8103 & \textbf{0.8382}  \\
\hline
\end{tabular}
\end{table}

\begin{table}
\centering
\caption{Nursing Dataset performance}
\label{tab:nursingresults}
\begin{tabular}{|l|l|ccc|ccc|} 
\hline
\multicolumn{1}{|c|}{\multirow{2}{*}{PHI Category}} & \multicolumn{1}{c|}{\multirow{2}{*}{PHI Sub-Type}} & \multicolumn{3}{c|}{ Burckhardt et al. } & \multicolumn{3}{c|}{Our architecture}      \\ 
\cline{3-8}
\multicolumn{1}{|c|}{}                              & \multicolumn{1}{c|}{}                              & Precision & Recall & $F_1$                 & Precision & Recall & $F_1$           \\ 
\hline
Name                                                & Patient Name                                       &           & 0.944  &                       & 0.866     & 0.591  & 0.703           \\
                                                    & Clinician Name                                     &           & 0.925  &                       & 0.936     & 0.830  & 0.880           \\
                                                    & Relative/Proxy Name                                &           & 0.989  &                       & 0.851     & 0.816  & 0.833           \\ 
\hline\hline
Name (overall)                                      &                                                    & 0.734     & 0.95   & 0.828                 & 0.982     & 0.866  & \textbf{0.920}  \\
Date                                                &                                                    & 0.256     & 0.992  & 0.407                 & 0.915     & 0.788  & \textbf{0.847}  \\
Location                                            &                                                    & 0.922     & 0.741  & 0.822                 & 0.963     & 0.839  & \textbf{0.897}  \\
Phone                                               &                                                    & 0.899     & 1.0    & \textbf{0.947}        & 0.778     & 0.583  & 0.667           \\
Overall                                             &                                                    & 0.645     & 0.919  & 0.758                 & 0.914     & 0.743  & \textbf{0.812}  \\
\hline
\end{tabular}
\end{table}
\section{Discussion and Error Analysis}
\label{sec:disAndFurther}

As we can see in Table \ref{tab:HIPAAphi}, with the exception of \textit{ID}, our architecture performs 
considerably better than systems by Liu et al. and Yang et al. Dernoncourt et
al. did not provide exact figures for the HIPAA-PHI categories so we have
excluded them from our analysis. Furthermore, Table \ref{tab:i2b2binary} shows
that our architecture performs similarly to the best scores achieved by
Dernoncourt et al., with our architecture slightly edging out Dernoncourt et al.
on the precision metric. For the nursing corpus, our system, while not performing as well as the
performances on i2b2 data set, managed to best the scores achieved by the
\textit{deidentify} system while also achieving a binary $F_1$ score of over
0.812. It is important to note that \textit{deidentify} was a
unsupervised learning system, it did not require the use of a train-valid-test
split and therefore, used the whole data set for their performance numbers.
The results of our architecture is assessed using a 60\%/40\% train/test split.

Our architecture noticeably converges faster than the NeuroNER, which was
trained for 100 epochs and the system by Liu et al.
\cite{harbin_deid_using_rnn_and_crf} which was trained for 80 epochs. 
Different runs of training our architecture on the i2b2 dataset converge at around 23$\pm$4 
epochs. A possible explanation for this is due to our architecture using the
Adam optimizer, whereas the NeuroNER system use the Stochastic Gradient Descent
(SGD) optimizer. In fact, Reimers et al.
\cite{reimers17_optim_hyper_deep_lstm_networ} show that the SGD optimizer performed
considerably worse than the Adam optimizer for different NLP tasks.

Furthermore, we also do not see any noticeable improvements from using the PubMed database
trained word embeddings \cite{bioEmbeddingSource} instead of the general text trained \textit{GloVe}
word embeddings. In fact, we consistently saw better $F_1$ scores using the \textit{GloVe} embeddings. 
This could be due to the fact that our use case was for identifying general labels such as Names, Phones, Locations etc.
instead of bio-medical specific terms such as diseases which are far better represented in the PubMed corpus.

\subsection{Error Analysis}
\label{sec:erroranalysis}

\subsubsection{i2b2 Dataset}
\label{sec:i2b2dataseterror}

We will mainly focus on the two PHI categories: \textit{Profession} and \textit{ID}
for our error analysis on the i2b2 data set. It is interesting to note that the
best performing models on the i2b2 data set by Dernoncourt et al.
\cite{MIT_dernoncourt} experienced similar lower performances on the same two
categories. However, we note the performances by Dernoncourt et al. were achieved using a ``combination of n-gram,
morphological, orthographic and gazetteer features'' \cite{MIT_dernoncourt}
while our architecture uses only POS tagging as an external feature. Dernoncourt
et al. posits that the lower performance on the \textit{Profession} category might be due
to the close embeddings of the \textit{Profession} tokens to other PHI tokens
which we can confirm on our architecture as well. Furthermore, our experiments
show that the \textit{Profession} PHI performs considerably better with the PubMed embedded
model than \textit{GloVe} embedded model. This could be due to the fact that
PubMed embeddings were trained on the PubMed database, which is a database of
medical literature. \textit{GloVe} on the other hand was trained on a general database, which
means the PubMed embeddings for \textit{Profession} tokens might not be as close
to other tokens as is the case for the \textit{GloVe} embeddings.

For the \textit{ID} PHI, our analysis shows that some of the errors were due to
tokenization errors. For example, a ``:'' was counted as PHI token which our
architecture correctly predicted as not a PHI token. Since our architecture is not custom
tailored to detect sophisticated ID patterns such as the systems in
\cite{MIT_dernoncourt, nottingham_i2b2_paper}, we have failed to detect some
\textit{ID} PHIs such as ``265-01-73'', a medical record number, which our
architecture predicted as a phone number due to the format of the number. Such
errors could easily be mitigated by the use of simple regular expressions.

\subsubsection{Nursing Dataset}
\label{sec:nursingdataerror}

We can see that our architecture outperforms the \textit{deidentify} system
by a considerable margin on most categories as measured by the $F_1$ score.
For example, the authors of \textit{deidentify} note that \textit{Date} PHIs
have considerably low precision values while our architecture
achieve a precision value of greater than 0.915\% for the \textit{Date} PHI.
However, Burckhardt et al. \cite{philip_deidentify} achieve an impressive
precision of 0.899 and recall of 1.0 for the \textit{Phone} PHI while our
architecture only manages 0.778 and 0.583 respectively. Our analysis of this
category shows that this is mainly due a difference in tokenization, stand alone
number are being classified as not a PHI. 

We tried to use the model that we trained on the i2b2 data set to predict the
categories of the nursing data set. However, due to difference in the text
structure, the actual text and the format, we achieved less than random
performance on the nursing data set. This brings up an important point about the
transferability of such models.

\subsection{Ablation Analysis}
\label{sec:ablation}

\begin{figure}[h]
 \centering
\caption{Ablation Analysis}
\label{fig:ablation}
\includegraphics[scale=0.6]{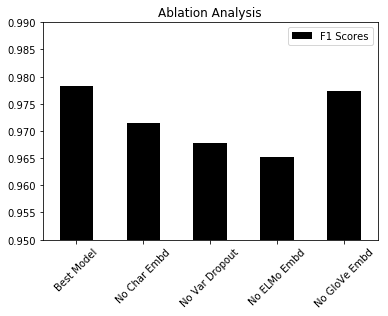}
\end{figure}

Our ablation analysis shows us that the layers of our models adds to the
overall performance. Figure \ref{fig:ablation} shows the binary $F_1$ scores on the i2b2 data set with each
bar being a feature toggled off. For example, the ``No Char Embd'' bar shows the
performance of the model with no character embeddings and everything else the
same as our best model.

We can see a noticeable change in the performance if we do not include the
\textit{ELMo} embeddings versus no \textit{GloVe} embeddings. The slight
decrease in performance when we use no \textit{GloVe} embeddings shows us that
this is a feature we might choose to exclude if computation time is limited.
Furthermore, we can see the impact of having no variational dropout and only
using a na\"ive dropout, it shows that variational dropout is better at
regularizing our neural network.

\section{Conclusion}
\label{sec:conclusion}

In this study, we show that our deep learning architecture, which incorporates the latest
developments in contextual word embeddings and NLP, achieves state-of-the-art
performance on two widely available gold standard de-identification data sets
while also achieving similar performance as the best system available in less
epochs. Our architecture also significantly improves over the performance of the hybrid system
\textit{deidentify} on the nursing data set.

This architecture could be integrated into a client-ready system such as the
\textit{deidentify} system. However, as mentioned in Section
\ref{sec:disAndFurther}, the use of a dictionary (or gazetter) might help
improve the model even further specially with regards to the \textit{Location}
and \textit{Profession} PHI types. Such a hybrid system would be highly
beneficial to practitioners that needs to de-identify patient data on a daily
basis.  

\bibliographystyle{ieeetr}
\bibliography{references}
\end{document}